\documentclass[conference]{IEEEtran}
\IEEEoverridecommandlockouts
\usepackage{cite}
\usepackage{amsmath,amssymb,amsfonts}
\usepackage{algorithmic}
\usepackage{array}
\usepackage{graphicx}
\usepackage{textcomp}
\usepackage{multirow}
\usepackage{xcolor}
\usepackage{blindtext}
\usepackage{fancyhdr}
\usepackage{csquotes}
\usepackage{tabularx}

\begin{document}

\title{CheXFuse: Guided Differential Joint Attention Fusion for Robust Multimodal Chest X-Ray Classification in Data-Scarce Medical Settings \\
}

\author{\IEEEauthorblockN{Muntaqim Ahmed Raju\IEEEauthorrefmark{1}, Priyanka Siddappa\IEEEauthorrefmark{1}, Md Shifat Haider Al Amin\IEEEauthorrefmark{1}\\
Ruizhe Ma\IEEEauthorrefmark{1}, and Md Ripon Ahammed\IEEEauthorrefmark{2}}
\IEEEauthorblockA{\IEEEauthorrefmark{1}University of Massachusetts Lowell, USA}
\IEEEauthorblockA{\IEEEauthorrefmark{2}NYC Health \& Hospitals/Jacobi-NCB, USA}
\IEEEauthorblockA{Email: \{MuntaqimAhmed\_Raju, Priyanka\_Siddappa, MdShifat\_AlAmin\}@student.uml.edu\\
Ruizhe\_Ma@uml.edu, Ripon.Ahammed90@gmail.com}
}

\maketitle

\begin{abstract}
A central challenge in multimodal learning is how to fuse modalities
that differ substantially in predictive strength. When one modality
provides a much stronger representation than another, indiscriminate
cross-modal interaction can introduce weak, redundant, or unstable
signals, limiting or even degrading performance relative to the
strongest unimodal representation. To address this problem, we propose
CheXFuse, a multimodal fusion framework based on Guided Differential
Joint Attention (GuDJA). CheXFuse combines image and text token
representations within a joint multimodal sequence, where learned token
guidance emphasizes informative representations and differential
attention reduces less useful attention responses. We evaluate
CheXFuse on the Indiana University (IU) Chest X-ray dataset, using
binary (normal vs. abnormal) classification as the downstream task.
Experiments are conducted under both original-report and masked-report
settings, with the latter serving as a controlled stress test under
reduced textual informativeness. CheXFuse with GuDJA achieves the
strongest mean overall performance across the evaluated baselines and
ablations, reaching 95.73\% accuracy, 95.79\% F1-score, 96.54\%
recall, and 95.28\% sensitivity at 95\% specificity (Sens@95Spec) with
original reports, and 95.58\%, 95.60\%, 95.60\%, and 94.34\%, respectively, under masked reports. These results suggest that GuDJA provides a promising fusion strategy under asymmetric or partially reduced modality informativeness.
\end{abstract}


\section{Introduction}

Multimodal learning aims to integrate information from heterogeneous
sources to construct representations that capture complementary
information across modalities. However, effective multimodal fusion
remains challenging because different modalities can exhibit distinct
representation structures, statistical properties, and learning
behaviors \cite{baltruvsaitis2018multimodal,ramachandram2017deep,raju2025multimodal}.
Importantly, access to multiple modalities does not necessarily
guarantee better predictive performance than the strongest unimodal
representation; prior work has shown that multimodal models can
underperform unimodal counterparts when the constituent modalities
differ in their learning and generalization behavior
\cite{wang2020makes}. These observations motivate fusion
mechanisms that can selectively integrate useful cross-modal
information while limiting the influence of weak or redundant
interactions.

Paired chest radiographs and radiology reports provide a particularly
relevant image--text setting for studying this problem. Public
radiology collections associate chest radiographs with free-text
reports written to document the radiographic examination
\cite{demner2016preparing,johnson2019mimiccxr}. The two
modalities encode information about the same clinical case in
fundamentally different forms: radiographs represent anatomical and
pathological patterns spatially, whereas reports provide contextual
language describing radiographic observations and their clinical
interpretation. This structural and semantic heterogeneity makes paired chest radiograph--report data particularly suitable for studying how visual and textual representations can be fused effectively.

Prior studies have explored chest radiograph--report learning through
attention-based interaction, semantic supervision, and multimodal
representation learning. TieNet jointly modeled visual and textual
information for thoracic disease classification and report generation
\cite{wang2018tienet}, while TandemNet used diagnostic reports as
semantic references for medical image classification
\cite{zhang2017tandemnet}. More recent approaches, including ConVIRT,
GLoRIA, BioViL, and MedKLIP, have demonstrated the value of paired
radiographs and reports for medical representation learning,
label-efficient classification, and image--text alignment
\cite{zhang2022convirt,huang2021gloria,boecking2022biovil,
wu2023medklip}. Together, these studies establish the value of paired
image--text information for learning radiographic representations and
motivate further investigation of how the two modalities should
interact during downstream multimodal learning.

A remaining challenge is that the two modalities may contribute very
differently to the prediction task. Radiology reports can contain
highly discriminative semantic information, whereas subtle visual
abnormalities may be more difficult for an image encoder to capture,
particularly when training data are limited. Consequently,
image--report learning can operate under strongly asymmetric modality
informativeness. In such settings, additional cross-modal interaction
does not necessarily improve the resulting representation and may
instead introduce weak or redundant information into an otherwise
strong unimodal representation. The problem becomes more difficult
when modality informativeness varies across samples or is partially reduced. Effective fusion therefore requires not only combining the
modalities, but also controlling which cross-modal interactions are
allowed to influence the final representation.

Attention-based fusion provides a natural framework for modeling such
interactions because it preserves token- or region-level structure. However, conventional attention typically determines interaction strength through normalized similarity scores and may therefore retain responses that are redundant or only weakly informative. When modality informativeness is uneven, such interactions can limit the benefit of multimodal fusion. This motivates a more selective mechanism that can emphasize useful representations before attention is computed while suppressing less informative attention responses during interaction.

In our prior work, guided feature emphasis was used to prioritize
task-relevant representations in medical imaging
\cite{raju2026cirrguide}. Building on this broader principle, we
combine learned token guidance with differential attention. Token
guidance modulates the representations before attention is computed,
enabling the model to emphasize image and text tokens that are more
useful for subsequent interaction. Differential attention provides a
complementary mechanism by forming two attention distributions and
subtracting one from the other, thereby offering a means of reducing
less informative responses \cite{ye2024differential}. The two
mechanisms are complementary: guidance controls which representations
enter the interaction, while differential attention controls which
attention responses are retained.

Based on this motivation, we propose CheXFuse, a multimodal fusion
framework based on Guided Differential Joint Attention (GuDJA).
CheXFuse uses a CheXNet-14 pretrained DenseNet121 encoder to obtain
spatial radiograph tokens and BioClinicalBERT to obtain contextual
report tokens. The two token sequences are combined within a joint
multimodal representation. GuDJA first applies learned guidance to
modulate the representations entering attention and then performs
differential attention to reduce less useful responses. We evaluate CheXFuse using paired chest radiographs and radiology
reports from the Indiana University (IU) Chest X-ray dataset, with binary (normal vs. abnormal) classification serving as the downstream
evaluation task. This data-scarce paired image--text setting exhibits
substantial modality asymmetry and therefore provides a challenging
setting for evaluating the proposed fusion mechanism. We consider two
complementary conditions. The original-report setting evaluates fusion
when the complete report is available, while a controlled masked-report
setting weakens explicit diagnostic keywords, negation expressions,
and label-associated textual shortcuts. The masked condition therefore
serves as a stress test in which the informativeness of the dominant
textual modality is partially reduced. These experiments are intended
to characterize the behavior of the proposed fusion mechanism under
different levels of modality informativeness.

The main contributions of this work are summarized as follows:

\begin{itemize}

\item We formulate paired chest radiograph--report classification as
a robust multimodal fusion problem under asymmetric modality
informativeness, where additional cross-modal interaction does not
necessarily improve upon a strong unimodal representation.

\item We propose CheXFuse with Guided Differential Joint Attention
(GuDJA), which combines learned token guidance with differential
attention to selectively strengthen informative joint interactions
while reducing less useful attention responses.

\item We introduce a controlled hybrid report-masking strategy that
weakens explicit diagnostic keywords, negation expressions, and
label-associated textual shortcuts to evaluate fusion robustness when
the informativeness of the dominant textual modality is partially
reduced.

\end{itemize}

\section{Method}

\begin{figure*}[t]
\centering
\includegraphics[width=.95\textwidth]{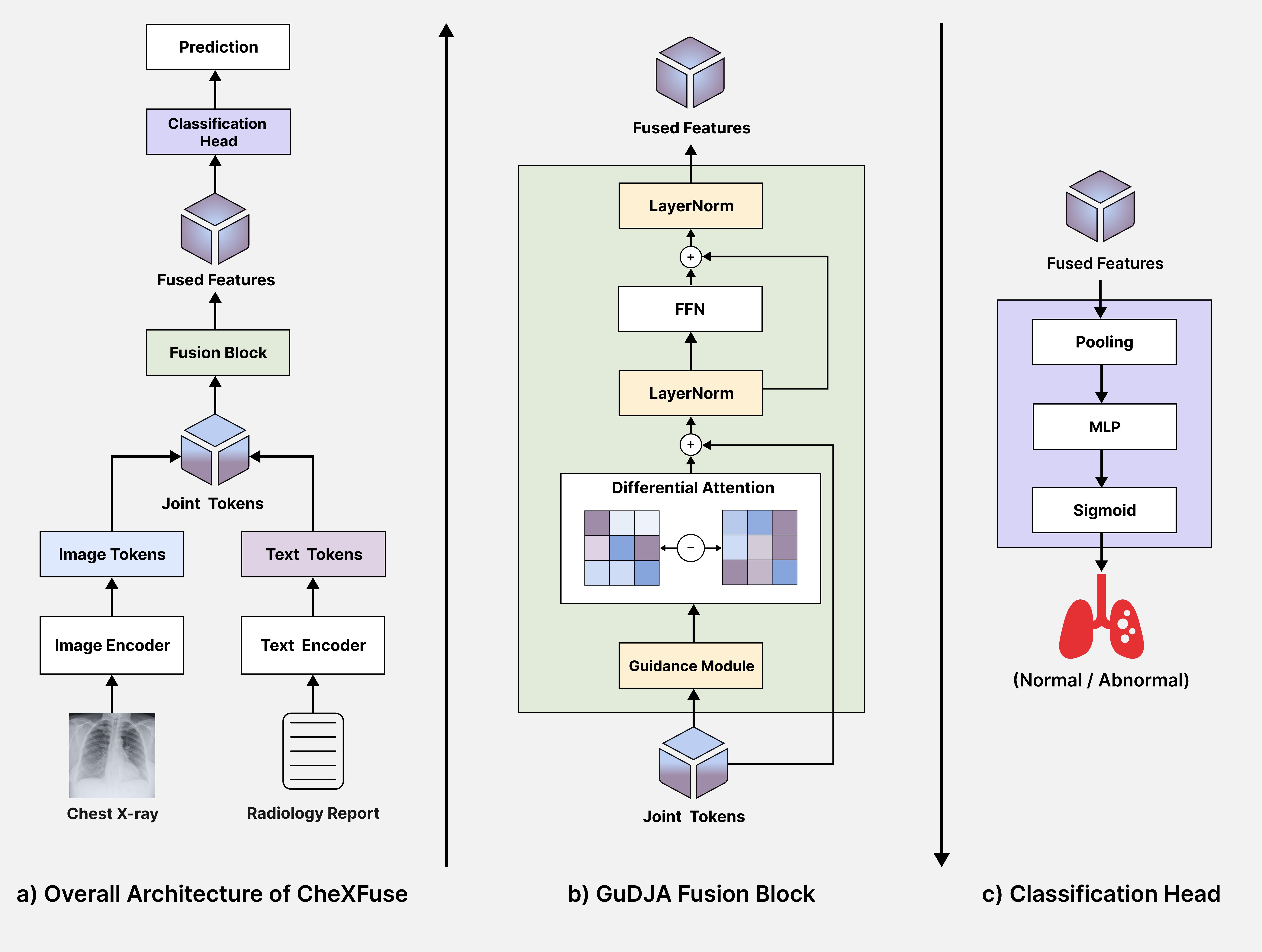}
\caption{Overview of the proposed CheXFuse framework, consisting of
separate image and text encoders, token-level multimodal fusion through
the GuDJA module, and a classification head for binary
(normal vs. abnormal) prediction.}
\label{fig}
\end{figure*}

This section describes CheXFuse, a multimodal fusion framework
evaluated on paired chest radiograph--report classification.
Fig.~\ref{fig} illustrates the overall architecture, which consists
of separate image and text encoders, token-level multimodal fusion
through the proposed GuDJA module, and a classification head for binary prediction.

Given a paired chest X-ray image and radiology report sample $(x_I, x_T, y)$, where $x_I$ denotes the input image, $x_T$ denotes the corresponding report text, and $y \in \{0,1\}$ is the binary class label, our objective is to predict whether the image-report pair corresponds to a normal or abnormal case. The overall fusion framework consists of three main components: image feature extraction, text feature extraction, and the proposed GuDJA fusion module.

First, we extract spatial visual tokens from the input chest X-ray image using a CheXNet-14 pretrained DenseNet121 encoder. Let $E_I(\cdot)$ denote the image encoder. Given an input image $x_I$, the convolutional backbone produces a spatial feature map:

\begin{equation}
F_I = E_I(x_I),
\end{equation}

where $F_I \in \mathbb{R}^{B \times C \times H' \times W'}$, $B$ is the batch size, $C$ is the number of feature channels, and $H'$ and $W'$ denote the height and width of the spatial feature map. The feature map is then projected into a shared multimodal embedding dimension $d$ using a $1 \times 1$ convolution:

\begin{equation}
\tilde{F}_I = P_I(F_I),
\end{equation}

where $\tilde{F}_I \in \mathbb{R}^{B \times d \times H' \times W'}$ and $P_I(\cdot)$ denotes the projection layer. The projected feature map is flattened along the spatial dimensions to obtain a sequence of image tokens:

\begin{equation}
H_I = \mathrm{Flatten}(\tilde{F}_I),
\end{equation}

where $H_I \in \mathbb{R}^{B \times N_I \times d}$ and $N_I$ is the number of image tokens. Layer normalization and learnable image positional embeddings are applied to preserve spatial information in the token representation:

\begin{equation}
H_I = \mathrm{LN}(H_I) + P_{\mathrm{pos}}^I.
\end{equation}

For textual representation, the radiology report $x_T$ is tokenized and passed through a BioClinicalBERT encoder. Let $E_T(\cdot)$ denote the text encoder and $m_T$ denote the text attention mask, where valid tokens are assigned 1 and padding tokens are assigned 0. The text encoder produces contextual token embeddings from the final hidden layer:

\begin{equation}
H_T = E_T(x_T, m_T),
\end{equation}

where $H_T \in \mathbb{R}^{B \times N_T \times d}$ and $N_T$ is the number of text tokens. 

Next, the image and text token sequences are concatenated to form a joint multimodal token representation:

\begin{equation}
H = \mathrm{concat}(H_I, H_T),
\end{equation}

where $H \in \mathbb{R}^{B \times N \times d}$ and $N = N_I + N_T$. Image tokens are treated as valid tokens, while text padding tokens are ignored using the text attention mask. This joint token sequence is then passed to the proposed GuDJA module.

The guided component learns token-wise attention masks before applying differential attention over the joint image-text sequence. Specifically, query and key guidance masks are computed as:

\begin{equation}
G_Q = \sigma(W_{gQ2}\delta(W_{gQ1}H)),
\end{equation}

\begin{equation}
G_K = \sigma(W_{gK2}\delta(W_{gK1}H)),
\end{equation}

where $\delta(\cdot)$ denotes the ReLU activation function and $\sigma(\cdot)$ denotes the sigmoid function. These masks are used to modulate the joint token sequence:

\begin{equation}
H_Q = H \odot G_Q,
\end{equation}

\begin{equation}
H_K = H \odot G_K, \quad H_V = H \odot G_K,
\end{equation}

where $\odot$ denotes element-wise multiplication. This guidance mechanism allows the model to emphasize informative image-text tokens while reducing the effect of less relevant token interactions.

After guidance, differential attention is applied to the guided joint representation. Instead of using a single attention map, differential attention computes two separate attention distributions and subtracts one from the other to suppress noisy or redundant interactions. Given the guided query, key, and value representations, the projections are computed as:

\begin{equation}
Q_1 = H_Q W_{Q_1}, \quad K_1 = H_K W_{K_1},
\end{equation}

\begin{equation}
Q_2 = H_Q W_{Q_2}, \quad K_2 = H_K W_{K_2},
\end{equation}

\begin{equation}
V = H_V W_V.
\end{equation}

The differential attention output is then defined as:

\begin{equation}
\begin{aligned}
\mathrm{DiffAttn}(H_Q,H_K,H_V) =
\Bigg[ &
\mathrm{softmax}
\left(
\frac{Q_1K_1^{T}}{\sqrt{d_h}}
\right) \\
&-
\lambda
\mathrm{softmax}
\left(
\frac{Q_2K_2^{T}}{\sqrt{d_h}}
\right)
\Bigg]V,
\end{aligned}
\end{equation}

where $d_h$ is the dimension of each attention head and $\lambda$ is a
learnable scalar controlling the contribution of the subtractive attention
term. In our implementation, the underlying learnable parameter is
initialized to $0.5$ and constrained using a sigmoid function,
i.e., $\lambda=\sigma(0.5)\approx0.622$ at initialization. The attention logits corresponding to padded text tokens are masked before the softmax operation, ensuring that padding tokens do not contribute to the attention weights.

The output of the GuDJA module is followed by a residual connection, layer normalization, and a feed-forward network:

\begin{equation}
\hat{H} = \mathrm{LN}(H + \mathrm{DiffAttn}(H_Q,H_K,H_V)),
\end{equation}

\begin{equation}
H_F = \mathrm{LN}(\hat{H} + \mathrm{FFN}(\hat{H})).
\end{equation}

Finally, a padding-aware mean pooling operation is applied over the valid joint tokens to obtain the final multimodal representation:

\begin{equation}
z = \frac{\sum_{i=1}^{N} M_i H_{F,i}}{\sum_{i=1}^{N} M_i},
\end{equation}

where $M_i$ indicates whether the $i$-th token is valid. The pooled representation $z$ is passed to a multilayer classification head:

\begin{equation}
\ell = f_c(z),
\end{equation}

where $\ell$ is the output logit. The final abnormality probability is computed using the sigmoid function:

\begin{equation}
\hat{y} = \sigma(\ell).
\end{equation}

Thus, CheXFuse performs token-level multimodal fusion by jointly
modeling spatial radiograph tokens and contextual report tokens within
a shared representation. GuDJA introduces two complementary
operations: learned token-level guidance selectively weights the
representations entering attention, while differential attention uses
a subtractive combination of two attention distributions to reduce
less useful attention responses. This design provides a more selective
joint interaction mechanism for settings in which image and text
modalities may differ substantially in predictive strength and
informativeness.

\section{Experimental Evaluation}
\subsection{Dataset}

We evaluate the proposed CheXFuse framework using the IU Chest X-ray dataset~\cite{demner2016preparing}, as it reflects the data-scarce medical setting. The dataset consists of 8,121 images and 3,996 associated radiology reports~\cite{demner2016preparing}. In this study, the report text is extracted from the \textit{FINDINGS} section, while the binary (normal vs. abnormal) label is derived from the \textit {MeSH major} field. Since the image encoder uses CheXNet-14 pretrained weights based on DenseNet121, we retain only frontal-view chest radiographs. This ensures consistency with the CheXNet-14 pretraining domain, which used over 100,000 frontal chest radiographs for 14-label thoracic disease classification~\cite{rajpurkar2017chexnet}. After filtering non-frontal and unmatched image-report pairs, we construct a balanced dataset for binary (normal vs. abnormal) classification. 
The final processed dataset contains 2,129 paired samples, where each sample consisted of one frontal chest radiograph, its corresponding radiology report, and a binary class label. Abnormal cases are encoded as 1, while normal cases are encoded as 0. The dataset is partitioned at the patient level using an approximate 70/20/10 train-validation-test split, resulting in 1,490 training samples, 428 validation samples, and 211 test samples. The training and validation sets are class-balanced, while the test set is approximately balanced, containing 50.24\% abnormal and 49.76\% normal cases.

\begin{table*}[!t]
\caption{Examples of original and masked radiology reports.}
\label{tab:masking_examples}
\centering

\begin{tabular*}{\textwidth}{
@{\extracolsep{\fill}}
>{\centering\arraybackslash}m{0.15\textwidth}
>{\centering\arraybackslash}m{0.08\textwidth}
>{\raggedright\arraybackslash}m{0.34\textwidth}
>{\raggedright\arraybackslash}m{0.34\textwidth}
@{}
}
\hline
\textbf{Chest X-ray}
& \textbf{Label}
& \textbf{Original Report}
& \textbf{Masked Report} \\
\hline

\includegraphics[width=0.95\linewidth]{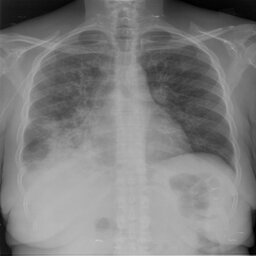}
& Abnormal
& There is diffuse right-sided airspace disease, with dense
consolidation in the right base. A right upper extremity PICC is seen
with the tip in the right brachiocephalic vein, representing an
interval retraction of approximately 6 cm. No pneumothorax or large
effusions. Heart size within normal limits.
& There is diffuse right-sided \texttt{[MASK]}, with \texttt{[MASK]}
in the right base. A right upper extremity PICC is seen with the tip
in the right brachiocephalic vein, representing an interval retraction
of approximately 6 cm. \texttt{[MASK]}. Heart size within
\texttt{[MASK]}. \\
\hline

\includegraphics[width=0.95\linewidth]{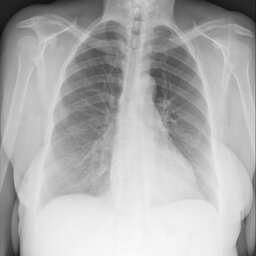}
& Normal
& The lungs are clear. There is no pleural effusion or pneumothorax.
The heart and mediastinum are normal. The skeletal structures are
normal.
& The lungs are \texttt{[MASK]}. There is \texttt{[MASK]}. The heart
and mediastinum are \texttt{[MASK]}. The skeletal structures are
\texttt{[MASK]}. \\
\hline

\end{tabular*}
\end{table*}

\subsection{Masked-Report Robustness Setting}
In addition to evaluation with the original report text, we introduce
a masked-report setting as a controlled stress test under reduced
textual informativeness. The objective is to examine how unimodal and
multimodal models behave when highly discriminative textual cues are
partially weakened. Explicit diagnostic keywords, negation phrases,
and label-associated shortcut expressions are therefore selectively
masked while preserving the remaining report context. This setting
does not remove the textual modality entirely; rather, it reduces some
of its most direct diagnostic signals and enables evaluation of fusion
robustness when the informativeness of the dominant textual modality
is partially reduced. The masked reports are generated from the original findings text using a hybrid masking strategy. First, rule-based diagnostic shortcut and
negation expressions are identified, including phrases such as
\enquote{within normal limits}, \enquote{lungs are clear},
\enquote{no pneumothorax}, and \enquote{no pleural effusion}.
Second, discriminative clinical keywords associated with normal and
abnormal reports are identified, including terms such as
\enquote{normal}, \enquote{clear}, \enquote{opacity},
\enquote{effusion}, \enquote{pneumothorax}, and
\enquote{cardiomegaly}. Third, data-driven phrase mining is performed
exclusively on the training set using document-level statistics over
1--4-grams. Candidate phrases occurring in at least five training
reports are ranked according to the magnitude of their smoothed
label-association score, and the top 60 phrases are retained. The
mined phrases are combined with the rule-based shortcut expressions,
and the resulting masking rules are then applied unchanged to the
validation and test sets. Selected spans are replaced with the \texttt{[MASK]} token. Negation and shortcut phrases are always masked, whereas individual
normality- and abnormality-associated clinical keywords are
independently masked with probability 0.5. A fixed random seed is used to ensure deterministic generation of the masked datasets. The resulting masked text is used as an alternative report input for evaluating model behavior when explicit textual
diagnostic information is partially reduced. Table~\ref{tab:masking_examples} shows examples of the masking strategy applied to radiology reports. The masked reports preserve the general sentence structure while reducing explicit diagnostic cues, negation expressions, and label-associated shortcut phrases.

\subsection{Experimental Setup}
All experiments are conducted using the processed IU Chest X-ray
dataset described above. Each sample consists of a frontal chest
radiograph, its corresponding radiology report, and a binary
(normal vs. abnormal) label. The original FINDINGS text is used in the original-report setting, while the corresponding masked report is used in the masked-report setting. The same training, validation, and test splits
are used across all evaluated models to ensure a consistent comparison.

For the image branch, we use a DenseNet121 encoder initialized with
CheXNet-14 pretrained weights. Because the pretrained backbone was
previously optimized on a large collection of frontal chest
radiographs, we keep the DenseNet121 feature extractor frozen and
train only the image-token projection layer together with the
subsequent multimodal fusion components. This fixed-feature strategy
preserves the pretrained radiographic representation while reducing
the number of trainable parameters in the data-scarce IU setting.
During training, images are transformed using random resized cropping
to $224 \times 224$ and horizontal flipping. For validation and
testing, images are resized to $256 \times 256$ and center-cropped to
$224 \times 224$. All images are normalized using the ImageNet mean
and standard deviation.

For the text branch, we use BioClinicalBERT to encode the radiology
reports. Each report is tokenized with a maximum sequence length of
128 tokens. Text sequences shorter than this length are padded, while
longer reports are truncated. The corresponding attention masks are
used during fusion to prevent padded tokens from contributing to
attention and pooling operations. Unlike the frozen image backbone,
the BioClinicalBERT encoder is fine-tuned during multimodal training.

All fusion models use a shared embedding dimension of $d=768$ and
8 attention heads, with a dropout rate of 0.2. The final classifier
consists of a fully connected layer, ReLU activation, dropout, and a
final linear layer that outputs a single logit for binary
classification.

The multimodal fusion models are trained using binary cross-entropy
with logits loss and optimized using AdamW with a learning rate of
$1 \times 10^{-4}$ and weight decay of $1 \times 10^{-4}$. A StepLR
scheduler is applied with a step size of 3 and decay factor
$\gamma=0.1$. The batch size is 16, and training proceeds for up to
10 epochs. Early stopping is based on validation accuracy with a
patience of 3 epochs and a minimum improvement threshold of
$1 \times 10^{-4}$.

For the image-only baseline, the same CheXNet-14 pretrained DenseNet121
backbone is used, but the backbone is fine-tuned jointly with the binary classification head. The model is optimized using AdamW with a learning rate of $1 \times 10^{-3}$ and
weight decay of $1 \times 10^{-4}$, while retaining the same batch
size, scheduler, maximum number of epochs, and early-stopping protocol
used for the multimodal experiments.

Each experiment is repeated using three random seeds: 7, 42, and 123
to account for variability due to random initialization and data
shuffling. For each seed, model selection is based on validation
accuracy, and the selected checkpoint is evaluated on the held-out
test set. Final results are reported as mean $\pm$ standard deviation
across the three runs.

\subsection{Evaluation Metrics}
We evaluate all models using accuracy, F1-score, recall, and
sensitivity at 95\% specificity (Sens@95Spec). Sens@95Spec measures
sensitivity using a decision threshold selected on the validation set
to correspond to the operating point whose specificity is closest to
95\%. For each run, this validation-selected threshold is then applied
unchanged to the held-out test set, and the resulting test sensitivity
is reported as Sens@95Spec.

For all reported results, the abnormal class is treated as the positive
class. Accordingly, recall and Sens@95Spec both quantify the ability to
correctly identify abnormal cases, but under different operating
conditions: recall is computed at the default classification threshold,
whereas Sens@95Spec is computed on the test set using the threshold
corresponding to specificity closest to 95\% on the validation set.

\subsection{Results and Discussion}

Tables~\ref{tab:original_results} and~\ref{tab:masked_results}
report the classification performance of the unimodal and multimodal
baselines, ablation variants, and the proposed CheXFuse with GuDJA
under the original-report and masked-report settings, respectively.
Results are reported as mean $\pm$ standard deviation over three
random seeds. We use these results to examine modality asymmetry, the
behavior of conventional fusion mechanisms, and the contribution of
guided differential joint attention under different levels of textual
informativeness.

\begin{table*}[!t]
\caption{Performance comparison using original radiology reports. Results are reported as mean $\pm$ standard deviation in percentage points over three random seeds (7, 42, and 123).}
\label{tab:original_results}
\centering
\begin{tabular}{llcccc}
\hline
Framework / Model
& Fusion Strategy
& Accuracy
& F1-score
& Recall
& Sens@95Spec \\
\hline

Unimodal (Image Only)\textsuperscript{a}
& --
& $70.46 \pm 0.72$
& $68.76 \pm 2.12$
& $65.09 \pm 7.72$
& $18.24 \pm 3.03$ \\

Unimodal (Text Only)\textsuperscript{a}
& --
& $95.26 \pm 0.07$
& $95.31 \pm 0.05$
& $95.91 \pm 1.09$
& $93.40 \pm 1.89$ \\

Multimodal\textsuperscript{a}
& Cross-Attention
& $95.10 \pm 1.52$
& $95.08 \pm 1.64$
& $94.65 \pm 3.57$
& $94.65 \pm 1.44$ \\

CheXFuse\textsuperscript{b}
& Joint Self-Attention
& $95.26 \pm 0.47$
& $95.26 \pm 0.52$
& $94.97 \pm 1.96$
& $91.82 \pm 1.44$ \\

CheXFuse\textsuperscript{b}
& Guided Joint Self-Attention
& $95.58 \pm 0.27$
& $95.63 \pm 0.26$
& $96.23 \pm 0.09$
& $94.34 \pm 2.50$ \\

CheXFuse\textsuperscript{b}
& Differential Joint Attention
& $95.58 \pm 0.27$
& $95.61 \pm 0.28$
& $95.91 \pm 0.54$
& $91.82 \pm 3.57$ \\

\textbf{CheXFuse}
& \textbf{GuDJA (Proposed)}
& $\mathbf{95.73 \pm 0.04}$
& $\mathbf{95.79 \pm 0.02}$
& $\mathbf{96.54 \pm 0.54}$
& $\mathbf{95.28 \pm 0.94}$ \\
\hline
\multicolumn{6}{l}{\footnotesize
\textsuperscript{a}Baseline; \textsuperscript{b}Ablation variant.}
\end{tabular}
\end{table*}

\begin{table*}[!t]
\caption{Performance comparison under the masked-report robustness setting. Results are reported as mean $\pm$ standard deviation in percentage points over three random seeds (7, 42, and 123).}
\label{tab:masked_results}
\centering
\begin{tabular}{llcccc}
\hline
Framework / Model
& Fusion Strategy
& Accuracy
& F1-score
& Recall
& Sens@95Spec \\
\hline

Unimodal (Text Only)\textsuperscript{a}
& --
& $94.94 \pm 0.27$
& $94.92 \pm 0.28$
& $94.03 \pm 0.54$
& $93.01 \pm 1.91$ \\

Multimodal\textsuperscript{a}
& Cross-Attention
& $94.63 \pm 0.27$
& $94.64 \pm 0.30$
& $94.34 \pm 0.94$
& $88.68 \pm 2.50$ \\

CheXFuse\textsuperscript{b}
& Joint Self-Attention
& $95.10 \pm 0.27$
& $95.12 \pm 0.29$
& $94.97 \pm 0.54$
& $83.33 \pm 15.82$ \\

CheXFuse\textsuperscript{b}
& Guided Joint Self-Attention
& $95.10 \pm 0.72$
& $95.11 \pm 0.76$
& $94.97 \pm 1.44$
& $93.08 \pm 1.96$ \\

CheXFuse\textsuperscript{b}
& Differential Joint Attention
& $94.94 \pm 0.72$
& $94.95 \pm 0.74$
& $94.65 \pm 1.44$
& $94.03 \pm 0.54$ \\

\textbf{CheXFuse}
& \textbf{GuDJA (Proposed)}
& $\mathbf{95.58 \pm 0.27}$
& $\mathbf{95.60 \pm 0.27}$
& $\mathbf{95.60 \pm 0.54}$
& $\mathbf{94.34 \pm 0.94}$ \\
\hline
\multicolumn{6}{l}{\footnotesize
\textsuperscript{a}Baseline; \textsuperscript{b}Ablation variant.}
\end{tabular}
\end{table*}

We first evaluated the relative predictive strength of the two
modalities using unimodal baselines. The image-only model achieved
$70.46 \pm 0.72\%$ accuracy, $65.09 \pm 7.72\%$ recall, and
$18.24 \pm 3.03\%$ Sens@95Spec. In contrast, the text-only model
achieved substantially stronger performance with original reports,
reaching $95.26 \pm 0.07\%$ accuracy, $95.91 \pm 1.09\%$ recall,
and $93.40 \pm 1.89\%$ Sens@95Spec. This large performance gap
indicates a strongly asymmetric multimodal setting in which the
radiology-report modality provides considerably more discriminative
information than the image-only representation on the processed IU
Chest X-ray dataset. Consequently, an effective multimodal fusion
method must not only incorporate complementary visual information,
but also avoid degrading an already strong textual representation
through weak or less informative cross-modal interactions.

When masked reports were used, the text-only model maintained high
overall accuracy at $94.94 \pm 0.27\%$, while recall decreased from
$95.91 \pm 1.09\%$ to $94.03 \pm 0.54\%$. Sens@95Spec remained
high at $93.01 \pm 1.91\%$. These results show that the report
representation remains strongly predictive even after explicit
diagnostic cues are partially reduced, although the decrease in
recall indicates some sensitivity to the removed textual information.
The masked-report setting therefore provides a controlled stress test
in which the informativeness of the dominant textual modality is
reduced without eliminating its remaining semantic context.

To examine whether conventional multimodal interaction can improve
upon this strong textual baseline, we next evaluated a text-to-image
cross-attention model. With original reports, cross-attention achieved
$94.65 \pm 1.44\%$ Sens@95Spec, numerically exceeding the
$93.40 \pm 1.89\%$ obtained by the text-only baseline at this
operating point. However, its overall performance did not consistently
exceed the text-only representation. More importantly, under the
masked-report setting, Sens@95Spec decreased to
$88.68 \pm 2.50\%$. This result illustrates that introducing
cross-modal interaction does not necessarily improve a strong unimodal
representation and may become less reliable when the informativeness
of the report modality changes.

A similar pattern was observed with standard joint self-attention.
With original reports, joint self-attention achieved
$95.26 \pm 0.47\%$ accuracy, comparable to the text-only baseline,
but its Sens@95Spec was lower at $91.82 \pm 1.44\%$. Under masked
reports, accuracy remained relatively high at $95.10 \pm 0.27\%$,
whereas Sens@95Spec decreased substantially to
$83.33 \pm 15.82\%$. The large standard deviation indicates
considerable seed-to-seed instability at this high-specificity
operating point. Together, the cross-attention and joint-attention
results demonstrate that simply introducing image-text interaction
does not guarantee more robust multimodal prediction, particularly
when the two modalities differ substantially in predictive strength.

The ablation variants further clarify the contributions of the two
components used in GuDJA. Guided joint self-attention achieved
$96.23 \pm 0.09\%$ recall and $94.34 \pm 2.50\%$ Sens@95Spec
with original reports, corresponding to mean gains of 1.26 and
2.52 percentage points, respectively, over standard joint
self-attention. Under masked reports, its Sens@95Spec increased from
$83.33 \pm 15.82\%$ for standard joint self-attention to
$93.08 \pm 1.96\%$. The substantially lower variability is
consistent with guided token weighting providing a more selective
interaction mechanism than conventional joint attention when textual
informativeness is reduced.

Differential joint attention showed a complementary pattern. With
original reports, it achieved $95.58 \pm 0.27\%$ accuracy and
$95.91 \pm 0.54\%$ recall, although Sens@95Spec remained at
$91.82 \pm 3.57\%$. Under masked reports, differential joint
attention achieved $94.03 \pm 0.54\%$ Sens@95Spec, representing
a 10.70 percentage-point increase over standard joint self-attention
together with substantially lower variability. This result is
consistent with the differential mechanism reducing less useful or
unstable attention responses when the information available to the
fusion module changes.

The proposed CheXFuse framework combines guided token weighting and
differential attention through GuDJA. With original reports, GuDJA
achieved the strongest mean overall performance among the evaluated
methods, obtaining $95.73 \pm 0.04\%$ accuracy,
$95.79 \pm 0.02\%$ F1-score, $96.54 \pm 0.54\%$ recall, and
$95.28 \pm 0.94\%$ Sens@95Spec. Relative to standard joint
self-attention, GuDJA improved mean recall by 1.57 percentage points
and Sens@95Spec by 3.46 percentage points. Importantly, GuDJA also
maintained the strong performance of the text-dominant representation
rather than exhibiting the degradation observed with several simpler
multimodal fusion strategies. Under the masked-report setting, GuDJA remained the strongest overall model, achieving $95.58 \pm 0.27\%$ accuracy,
$95.60 \pm 0.27\%$ F1-score, $95.60 \pm 0.54\%$ recall, and
$94.34 \pm 0.94\%$ Sens@95Spec. Relative to standard joint
self-attention, GuDJA improved mean Sens@95Spec by 11.01 percentage
points and showed substantially lower variability across seeds.
Compared with the masked text-only baseline, GuDJA improved mean
recall from $94.03 \pm 0.54\%$ to $95.60 \pm 0.54\%$, while
Sens@95Spec increased from $93.01 \pm 1.91\%$ to
$94.34 \pm 0.94\%$. Relative to cross-attention, GuDJA improved
masked-report recall by 1.26 percentage points and Sens@95Spec by
5.66 percentage points.

Taken together, these results characterize the central fusion challenge
considered in this study: the image and text modalities differ
substantially in standalone predictive strength, and conventional
cross-attention and joint self-attention show that additional
multimodal interaction does not necessarily improve upon a strong
unimodal representation. In contrast, GuDJA maintains strong
performance across both original-report and masked-report conditions,
with only a 0.94 percentage-point decrease in mean recall after
masking, compared with a 1.88 percentage-point decrease for the
text-only baseline. These findings indicate that combining guided
token weighting with differential attention provides a more robust
mechanism for multimodal interaction when modality informativeness is
asymmetric or partially reduced.

Paired exact McNemar tests comparing GuDJA with the text-only baseline did not reach statistical significance at the individual-seed level ($p>0.05$ for all comparisons). Accordingly, the improvements reported above should be interpreted as consistent numerical gains across the evaluated seeds rather than statistically significant superiority on the 211-case test set. Nevertheless, across both the original-report and masked-report settings, GuDJA consistently maintained strong overall and high-specificity classification performance while avoiding the larger degradation observed with several conventional multimodal fusion strategies when textual informativeness was reduced.

\subsection{Case-Level Modality Disagreement Analysis}

To further examine how GuDJA behaves when the two modalities provide
different predictions, we performed a case-level comparison of the
image-only, masked text-only, and GuDJA models on the same test cases.
The image-only and masked text-only models disagreed on 56, 59, and
54 cases for seeds 7, 42, and 123, respectively. Among these cases,
GuDJA correctly classified 53/56 (94.64\%), 57/59 (96.61\%), and
52/54 (96.30\%) cases, respectively.

Across the three seeds, 38 test cases exhibited persistent disagreement,
where the image-only and text-only models produced opposing predictions
for all three runs. GuDJA correctly classified 37 of these 38 cases
(97.37\%) across all three seeds. Among the persistent disagreements,
23 were abnormal cases that the image-only model consistently predicted
as normal while the masked text-only model correctly identified as
abnormal; GuDJA correctly classified all 23 cases. Conversely, 14 were
normal cases that the image-only model consistently predicted as
abnormal while the text-only model correctly classified as normal;
GuDJA also correctly classified all 14 cases. One persistent abnormal
case was correctly classified by the image-only model but classified as
normal by both the text-only model and GuDJA. These results indicate
that GuDJA handles most persistent unimodal conflicts reliably in this
test set, while the remaining failure case suggests that the fused
representation can still be influenced by the stronger textual modality
when image and report evidence diverge.

\section{Limitations and Future Work}
CheXFuse equipped with GuDJA demonstrates promising performance as a
multimodal fusion framework, but several limitations remain. First, this study focuses on the processed IU Chest X-ray dataset, a relatively data-scarce paired image--report setting. Evaluation on larger and more diverse datasets, such as MIMIC-CXR, is needed to assess generalizability across patient populations, reporting styles, and imaging conditions. Second, the present study considers binary (normal vs. abnormal) classification. Extending CheXFuse to multi-label disease classification will help determine whether the proposed fusion strategy remains effective for more complex multimodal prediction tasks. Third, the multimodal setting assumes that the radiology report is available together with the corresponding chest radiograph at inference time. Because the FINDINGS section is produced following radiographic interpretation, it can contain highly discriminative information related to the classification target. The present experiments should therefore be interpreted primarily as an evaluation of multimodal fusion under asymmetric modality informativeness rather than as a simulation of prospective pre-interpretation screening. Future work will consider settings in which the textual modality is less directly associated with the prediction target. Finally, the results reveal substantial asymmetry in modality strength, with the report representation considerably outperforming the image-only representation. Although GuDJA remains robust under this condition, it does not explicitly estimate modality reliability or relative contribution. Future work will investigate reliability- and conflict-aware fusion mechanisms, including adaptive modality weighting, uncertainty-aware fusion, modality dropout, and regularization against excessive modality dominance.

\section{Conclusion}
This study presented CheXFuse, a multimodal fusion framework that
integrates radiograph and radiology-report representations through
Guided Differential Joint Attention (GuDJA). The method addresses
multimodal learning under asymmetric modality informativeness, where
additional cross-modal interaction does not necessarily improve upon a
strong unimodal representation. GuDJA combines guided token weighting
with differential attention to promote informative interactions while
reducing less useful or unstable attention responses. Across the evaluated baselines and ablation variants, CheXFuse with GuDJA achieved the strongest mean overall performance under both original-report and masked-report conditions. The masked-report experiments further showed that GuDJA maintained strong performance
when explicit textual diagnostic information was partially reduced,
while several conventional fusion strategies exhibited greater
degradation or variability. These results support GuDJA as a promising
approach for robust multimodal fusion when modality informativeness is
asymmetric or partially reduced. The present study should be viewed primarily as a controlled evaluation of multimodal fusion using paired chest radiographs and  reports. Future work will evaluate the framework on larger and more
diverse datasets, extend it to multi-label prediction tasks, and explore
reliability-aware and conflict-aware fusion mechanisms for handling
case-specific modality imbalance and conflicting multimodal evidence.

\bibliographystyle{IEEEtran}
\bibliography{references}

\end{document}